\documentclass[review]{elsarticle}

\usepackage{lineno,hyperref}
\modulolinenumbers[5]

\usepackage[dvipsnames]{xcolor}

\usepackage{graphicx}
\usepackage{times}
\usepackage{epsfig}
\usepackage{graphicx}
\usepackage{amsmath}
\usepackage{amssymb}
\usepackage{pifont}
\usepackage{float}
\usepackage{xcolor}
\usepackage{bm}
\usepackage{times}
\usepackage{epsfig}
\usepackage{graphicx}
\usepackage{float}
\floatstyle{plaintop}
\restylefloat{table}
\usepackage{multirow}
\newcommand{\cmark}{\ding{51}}%
\usepackage{array}
\newcolumntype{x}[1]{>{\centering\arraybackslash\hspace{0pt}}p{#1}}
\usepackage{subcaption}
\usepackage[symbol]{footmisc}
\usepackage{soul,color}
\usepackage{graphicx}
\usepackage{times}
\usepackage{epsfig}
\usepackage{graphicx}
\usepackage{amsmath}
\usepackage{amssymb}
\usepackage{pifont}
\usepackage{float}
\usepackage{xcolor}
\usepackage{bm}
\usepackage{times}
\usepackage{epsfig}
\usepackage{graphicx}
\usepackage{float}
\floatstyle{plaintop}
\restylefloat{table}
\usepackage{multirow}
\usepackage{array}
\newcolumntype{x}[1]{>{\centering\arraybackslash\hspace{0pt}}p{#1}}
\usepackage{subcaption}
\usepackage[symbol]{footmisc}
\usepackage{wrapfig}
\usepackage{booktabs}
\usepackage{stfloats, caption}%

\usepackage{times}
\usepackage{epsfig}
\usepackage{graphicx}
\usepackage{amsmath}
\usepackage{amssymb}
\usepackage{footnote}
\usepackage{dsfont}

\usepackage{comment}
\usepackage{blindtext}
\usepackage{morewrites}
\usepackage{makecell}
\usepackage{pifont}%
\usepackage{multirow}
\usepackage{soul}

\soulregister\cite7
\soulregister\ref7
\soulregister\pageref7

\begin{document}

\begin{frontmatter}

\title{Cross-view Action Recognition Understanding \\ From Exocentric to Egocentric Perspective}

\author{Thanh-Dat Truong and Khoa Luu}
\address{\small Computer Vision and Image Understanding Lab\\
University of Arkansas\\
Fayetteville, 72701, USA}
\address{\texttt{\{tt032, khoaluu\}@uark.edu}}

\begin{abstract}
Understanding action recognition in egocentric videos has emerged as a vital research topic with numerous practical applications. With the limitation in the scale of egocentric data collection, learning robust deep learning-based action recognition models remains difficult. Transferring knowledge learned from the large-scale exocentric data to the egocentric data is challenging due to the difference in videos across views. Our work introduces a novel cross-view learning approach to action recognition (CVAR) that effectively transfers knowledge from the exocentric to the selfish view. First, we present a novel geometric-based constraint into the self-attention mechanism in Transformer based on analyzing the camera positions between two views. Then, we propose a new cross-view self-attention loss learned on unpaired cross-view data to enforce the self-attention mechanism learning to transfer knowledge across views.  Finally, to further improve the performance of our cross-view learning approach, we present the metrics to measure the correlations in videos and attention maps effectively. Experimental results on standard egocentric action recognition benchmarks, i.e., Charades-Ego, EPIC-Kitchens-55, and EPIC-Kitchens-100, have shown our approach's effectiveness and state-of-the-art performance.
\end{abstract}

\begin{keyword}
Cross-View Action Recognition; Self-Attention; Egocentric Action Recognition;
\end{keyword}

\end{frontmatter}

% \linenumbers

\section{Introduction}

\begin{figure}
    \centering
    \includegraphics[width=1.0\textwidth]{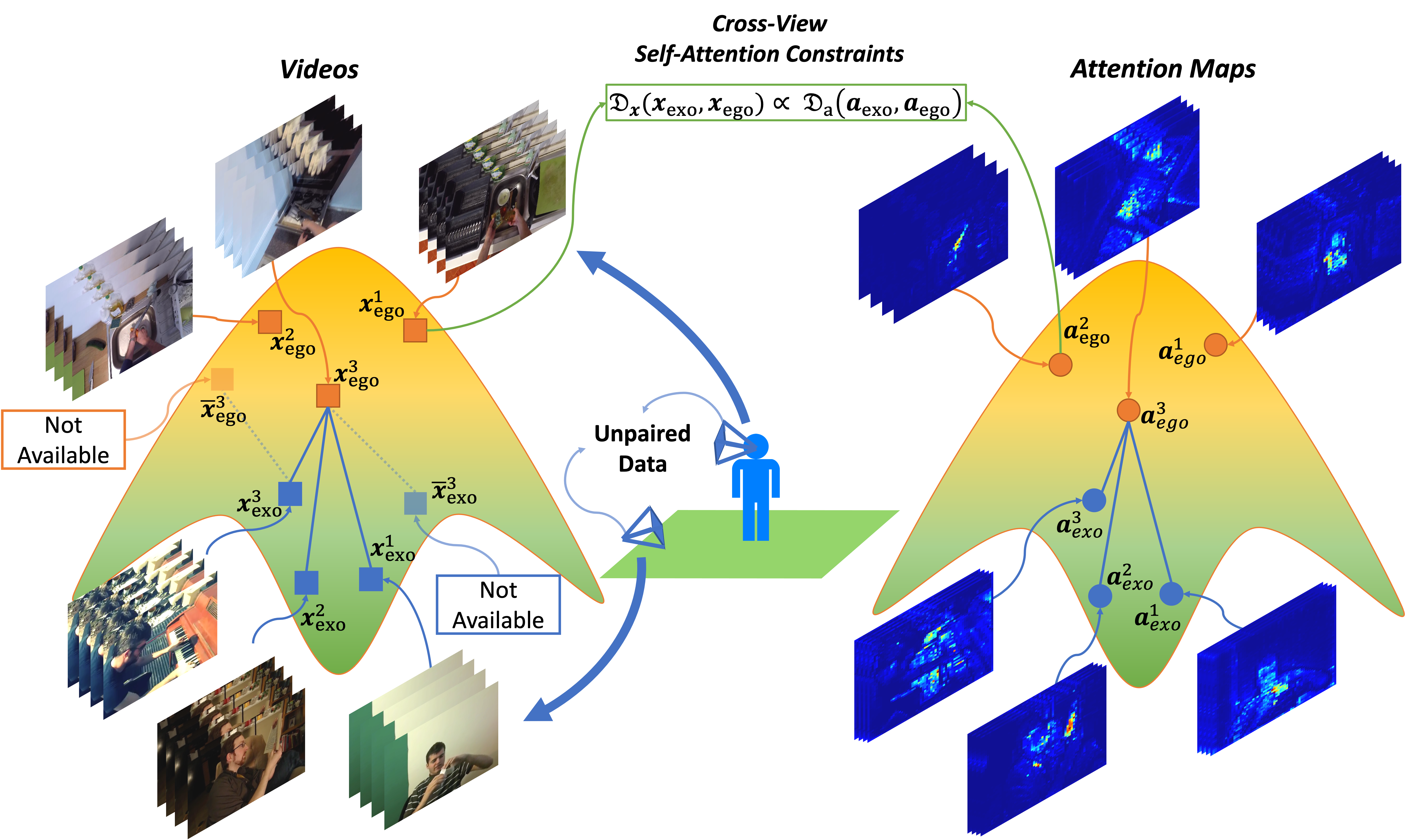}
    \caption{\textbf{The Cross-view Self-attention Constraints}. 
    Although under the setting of cross-view unpaired data where the corresponding video and its attention in the opposite view are inaccessible, our cross-view self-attention loss is proven to impose the cross-view constraints via unpaired samples based on the geometric properties between two camera positions.}
    \label{fig:figure_one}
\end{figure}

Analyzing first-view videos, i.e., egocentric videos, captured by wearable cameras has become an active research topic in recent years. With the recent development of virtual and augmented reality technologies,  this topic has gained more attention in the research communities due to the enormous interest in analyzing human behaviors from the first-view perspective.
Many tasks have been currently explored in the egocentric video data that provide many practical applications, e.g., action recognition \cite{ego-exo, swin, MViTv2, 9706375, egovlp}, action detection \cite{epic-100, untrimmednets, https://doi.org/10.48550/arxiv.2301.01380, Herzig_2022_CVPR, Soucek_2022_CVPR}, action anticipation \cite{egovlp, antonino-next-active, Liu_2022_CVPR}, etc.
In comparison with third-view video data, i.e., exocentric videos, egocentric videos provide new, distinct viewpoints of surrounding scenes and actions driven by the camera position holding on the observer.

The properties of egocentric videos also bring new challenges to video analysis tasks. One of the main issues is the scale of datasets. It is well known that learning robust video models, e.g., action recognition models, usually requires a large amount of video data \cite{swin, MViTv2, egovlp}. For example, the third-view action models are learned on the large-scale Kinetics-700 \cite{kinetic700} data comprising 650K videos over 700 classes. Meanwhile, the scale of egocentric video data is relatively small compared to third-view datasets, e.g., EPIC Kitchens \cite{epic-100} only consists of 90K clips or Charades-Ego \cite{charades-ego}  includes 68K clips of 157 classes. 
In addition, the egocentric video data lack variation, e.g., videos only in kitchens \cite{epic-100} or daily indoor activities \cite{charades-ego}. These problems pose a considerable challenge for learning robust video models on the first-view data.

Many prior works \cite{egovlp, slowfast}  have improved the performance of action recognition models by adopting the pre-trained model on large-scale third-view datasets and fine-tuning it on the first-view dataset. However, these methods often ignore the unique characteristics of egocentric videos. Thus, they could meet the unaligned domain problems. Another method \cite{ego-exo} has tried to alleviate this domain mismatch problem by introducing several additional egocentric tasks during the pre-training phase on the third-view datasets. However, this approach requires the labels of egocentric tasks on third-view data or relies on the off-the-shelf specific-task models. Domain adaptation methods \cite{ego-exo, ganin2016domain, sigurdsson2018actor} have also been utilized to transfer the knowledge from the third-view to first-view data. Nevertheless, these methods still need to model the camera-view changes during the adaptation phase.

With the recent success of Vision Transformer, the self-attention mechanism is fundamental to building an efficient action recognition model. Still, fewer prior works have focused on leveraging self-attention to model action recognition from the third-view to first-view data.
Moreover, modeling the change in camera positions across views is also one of the primary factors in sufficiently developing a learning approach from the exocentric to egocentric view. 
Therefore, considering these characteristics, we introduce a novel cross-view learning approach to effectively model the self-attention mechanism to transfer the knowledge learned on third-view to first-view data.
Our proposed approach first considers the geometric correlation between two camera views. Then, the cross-view geometric correlation constraint is further embedded into the self-attention mechanism so that the model can generalize well from the exocentric to the egocentric domain. Fig. \ref{fig:figure_one} illustrates the cross-view self-attention constraint.

\textbf{Contributions of this Work: } 
This work introduces a novel \textbf{C}ross-\textbf{V}iew learning approach to \textbf{A}ction \textbf{R}ecognition (CVAR) via effectively transferring knowledge from the exocentric video domain to the egocentric one. By analyzing the role of the self-attention mechanism and the change of camera position across views, we introduce a new geometric cross-view constraint for correlation between videos and attention maps. Then, from the proposed cross-view restriction, we present a novel cross-view self-attention loss that models the cross-view learning into the self-attention mechanism. Our proposed loss allows the Transformer-based model to adapt knowledge and generalize from the third-view to first-view video data. The cross-view correlations of videos and attention maps are further enhanced using the deep metric and the Jensen-Shannon divergence metric, respectively, that capture deep semantic information.
Our experimental results on the standard egocentric benchmark, i.e., Charades-Ego, EPIC-Kitchens-55, and EPIC-Kitchens-100, have illustrated the effectiveness of our proposed method and achieved state-of-the-art (SOTA) results.

\section{Related Work}

\noindent
\textbf{Video Action Recognition } 
Many large-scale third-view datasets have been introduced for action recognition tasks, e.g., Kinetics \cite{kinetic700, kinetic400, kinetic600}, Something-Something V2 \cite{ssv2}, Sport1M \cite{sport1m}, AVA \cite{ava}, etc. 
Many deep learning approaches \cite{swin, MViTv2, slowfast, i3d, tsn, lin2019tsm, vivit, mvit, truong2021direcformer} have been introduced and achieved remarkable achievements.
The early deep learning approaches \cite{sport1m, lrcn} have utilized the 2D Convolutional Neural Networks (CNNs) \cite{resnet, vgg, inceptionv3} to extract the deep spatial representation followed by using Recurrent Neural Networks (RNNs) \cite{lstm} to learn the temporal information from these extracted spatial features.
Some later approaches have improved the temporal learning capability by introducing the two-stream networks \cite{tsn, 2streams_simonyan, 2streams_feichtenhofer, spatiotemporal_resnet, spatiotemporal_multiplier} using both RGB video inputs and optical flows for motion modeling. 
Later, the 3D CNN-based approaches \cite{c3d} and their variants \cite{i3d, trn} have been introduced, i.e., 
several (2+1)D CNN architectures have been proposed \cite{slowfast, r21d, x3d, s3d}. Meanwhile, other approaches have used
pseudo-3D CNNs built based on 2D CNNs \cite{lin2019tsm, p3d}. In addition, to better capture the long-range temporal dependencies among video frames, the non-local operation has also been introduced \cite{non_local}.
SlowFast \cite{slowfast} proposes a dual-path network to learn spatiotemporal information at two different temporal rates. X3D \cite{x3d} progressively expands the networks to search for an optimal network for action recognition.

\noindent
\textbf{Vision Transformer} \cite{swin, MViTv2, vivit, mvit, vit, timesformer, truong2021direcformer, nguyen2024insect, nguyen2023micron} has become a dominant backbone in various tasks due to its outstanding performance. The early success of Video Vision Transformer (ViViT) \cite{vivit} has shown its promising capability in handling spatial-temporal tokens in action recognition.
Then, many variants \cite{swin, MViTv2, mvit, timesformer, truong2021direcformer, nguyen2024multi, nguyen2023hig} of ViViT have been introduced to improve the accuracy and reduce the computational cost. 
\cite{bulat2021spacetime} presented a space-time mixing attention mechanism to reduce the complexity of the self-attention layers.
TimeSFormer \cite{timesformer} introduced divided spatial and temporal attention to reduce the computational overhead. Then, it is further improved using the directed attention mechanism \cite{truong2021direcformer}.
Then, \cite{mvit} proposed a Multi-scale Vision Transformer (MViT) using multiscale feature hierarchies. Then, MViT-V2 \cite{MViTv2} improves the performance of MViT by incorporating decomposed relative positional embeddings and residual pooling connections.
Swin Video Transformer \cite{swin} has achieved state-of-the-art performance in action recognition by using shifted windows to limit the self-attention computational cost to local windows and also allow learning attention across windows.
The recent SVFormer \cite{xing2023svformer} has introduced a temporal warping augmentation to capture the complex temporal variation in videos for semi-supervised action recognition.
Meanwhile, MTV \cite{yan2022multiview} presented a Multiview Transformer to model different views of the videos with lateral connections to fuse information across views.
Later, M\&M Mix\cite{xiong2022m} further improved MTV by using multimodal inputs.
TADA \cite{huang2021tada} proposed Temporally-Adaptive Convolutions (TAdaConv) to model complex temporal dynamics in videos.
Inspired by the success of CLIP in vision-language pretraining \cite{radford2021learning, nguyen2023insect, jia2021scaling}, several studies have adopted this approach to video-language pretraining \cite{egovlp, wang2023all, sun2022long}.
All-in-One \cite{wang2023all} presented a unified approach to video-language pretraining by embedding raw video and textual inputs into joint representations
with a unified network.
EgoVLP \cite{egovlp} introduced Video-Language Pretraining to ego-centric video understanding. 
LF-VILA \cite{sun2022long} presented a Multimodal Temporal Contrastive and Hierarchical
Temporal Window Attention to model the long-form videos for video-language pretraining.

\noindent
\textbf{Egocentric Video Analysis}
Apart from third-view videos, egocentric videos provide distinguished viewpoints that pose several challenges in action recognition. Many datasets have been introduced to support the egocentric video analysis tasks, e.g., Charades-Ego \cite{charades-ego}, EPIC Kitchens \cite{epic, epic-100}, Ego4D \cite{Ego4D2022CVPR}, EgoClip \cite{egovlp}, HOI4D \cite{Liu_2022_CVPR_HOI4D}. These datasets provide several standard egocentric benchmarks, e.g., action recognition \cite{charades-ego, epic, Ego4D2022CVPR}, action anticipation \cite{Ego4D2022CVPR, epic-100}, action detection \cite{epic-100}, video-text retrieval \cite{egovlp, Ego4D2022CVPR}.
Many methods have been proposed for egocentric action recognition, including Multi-stream Networks \cite{ma2016going,li2018eye,epic-fusion,wang2020makes}, RNNs \cite{furnari2019would,furnari2020rolling,sudhakaran2019lsta}, 3D CNNs \cite{pirri2019anticipation,lu2019learning}, Graph Neural Networks \cite{ego-topo}.
Despite the difference in network designs, these prior works are usually pre-trained on the large-scale third-view datasets before fine-tuning them on the first-view dataset. However, there is a significant difference between the first-view and third-view datasets.
Thus, a direct fine-tuning approach without consideration of modeling view changes could result in less efficiency.
Many methods have improved the performance of the action recognition models by using additional egocentric cues or tasks, including gaze and motor attention \cite{mathe2012dynamic,li2018eye,liu2019forecasting}, object detection \cite{antonino-next-active,baradel2018object,dessalene2020egocentric,wang2020symbiotic}, hand interactions \cite{tekin2019h,shan2020understanding,kapidis2019egocentric}.
Ego-Exo \cite{ego-exo} presented an approach by introducing the auxiliary egocentric tasks into the pre-training phase on the third-view dataset, i.e., ego-score, object-score, and interaction map predictions.
However, these methods usually require the labels of auxiliary egocentric tasks on the third-view datasets or rely on pseudo labels produced by the off-the-shelf pre-trained models on egocentric tasks.

\noindent
\textbf{Cross-view Video Learning}
The cross-view learning approaches have been exploited and proposed for several tasks, e.g., geo-localization \cite{zhu2022transgeo, Toker_2021_CVPR, shi2020where, NIPS2019_9199}, semantic segmentation \cite{8885955, 9878624, DIMAURO2020175, truong2023falcon, truong2023fairness, truong2024conda}. Meanwhile, in video understanding tasks, several prior methods have alleviated the cross-view gap between exocentric and egocentric domains by using domain adaptation \cite{ganin2016domain, choi2020unsupervised, truong2023crovia, truong2024eagle}, learning viewpoint-invariant \cite{sigurdsson2018actor,soran2014action,ardeshir2018exocentric, shang2022learning},  or learning joint embedding \cite{xu2018joint, ardeshir2016ego2top, ardeshir2018integrating,ardeshir2016egoreid,yang2018ego}. 
Other works utilized generative models to synthesize the different viewpoints from a given image/video \cite{elfeki2018third,regmi2018cross,regmi2019bridging,liu2020exocentric}.
However, these methods often require either a pair of data of both first and third views to learn the joint embedding or a share label domain when using domain adaptation~\cite{truong2023fredom, truong2021bimal, truong2022otadapt}.

\section{Cross-view Learning in Action Recognition}

Let $\mathbf{x}_{exo} \in \mathbb{R}^{T \times H \times W \times 3}$ be a third-view (exocentric) video and $\mathbf{y}_{exo} \in \mathcal{Y}_{exo}$ be its corresponding ground-truth class, $\mathcal{Y}_{exo}$ is the set of classes in the exocentric dataset. Similarly, $\mathbf{x}_{ego} \in \mathbb{R}^{T \times H \times W \times 3}$ be a first-view (egocentric) video and $\mathbf{y}_{ego} \in \mathcal{Y}_{ego}$ be its corresponding ground-truth class, $\mathcal{Y}_{exo}$ is the set of classes in the egocentric dataset. Let $F: \mathbb{R}^{T \times H \times W \times 3} \to \mathbb{R}^{D}$ be the backbone network that maps a video into the deep representation, $C_{exo}$ and $C_{ego}$ are the classifier of exocentric and egocentric videos that predict the class probability from the deep representation. Then, the basic learning approach to learning the action model from the exocentric to the egocentric view can be formulated as a supervised objective, as in Eqn. \eqref{eqn:general_opt}.
\begin{equation} \label{eqn:general_opt}
\small
\begin{split}
    \arg\min_{\theta_F, \theta_{C_{exo}}, \theta_{C_{ego}}}[\mathbb{E}_{\mathbf{x}_{exo}, \mathbf{y}_{ego}}\mathcal{L}_{ce}(C_{exo}(F(\mathbf{x}_{exo})), \mathbf{y}_{exo}) \\ 
    + \mathbb{E}_{\mathbf{x}_{ego}, \mathbf{y}_{ego}}\mathcal{L}_{ce}(C_{ego}(F(\mathbf{x}_{ego})), \mathbf{y}_{ego})]
\end{split}
\end{equation}
where $\theta_F, \theta_{C_{exo}}, \theta_{C_{ego}}$ are the network parameters, $\mathcal{L}_{ce}$ is the supervised loss (i.e., cross-entropy loss). Several prior approaches \cite{ego-exo, sigurdsson2018actor} have adopted this learning approach to learn a cross-view action recognition model. Then, other prior methods have further improved the performance of models by using a large pretrained model  \cite{swin, slowfast}, 
domain adaptation  \cite{ganin2016domain}, learning a joint embedding between two views \cite{sigurdsson2018actor}, learning auxiliary egocentric tasks  \cite{ego-exo}.

Although these prior approaches \cite{ego-exo, swin, MViTv2, slowfast} showed their potential in improving performance, they have not effectively addressed the problem of cross-view learning. In particular, domain adaptation methods \cite{ganin2016domain} are often employed in the context of environment changes (e.g., simulation to real data), and the camera views are assumed on the same position (either third view or first view). However, there is a vast difference in videos between the third view and the first view. Thus, domain adaptation is considered less effective in the cross-view setting.
Meanwhile, fine-tuning the first-view action model on the large pretrained models \cite{swin, slowfast} usually relies on the deep representation learned from the large-scale third-view data. However, these deep representations do not have any guarantee mechanism well generalized in the first-view video. Also, learning the join embedding or auxiliary egocentric tasks \cite{ego-exo} suffer a similar problem due to their design of learning approaches without considering camera changes. In addition, it requires a pair of views of video data during training.
Therefore, to effectively learn the cross-view action recognition model, the learning approach should consider the following properties:
(1) the geometric correlation between the third view and the first view has to be considered during the learning process,
(2) the mechanism that guarantees the knowledge learned is well generalized from the third view to the first view.

\subsection{Cross-view Geometric Correlation in Attentions}

With the success of Vision Transformer in action recognition \cite{MViTv2, swin, vit},  the self-attention mechanism is the key to learning the robust action recognition models. Therefore, our work proposes explicitly modeling cross-view learning in action recognition models through the self-attention mechanism. First, we revise the geometric correlation of the exocentric and egocentric views in obtaining the videos. Let us assume that $\mathbf{\bar{x}}_{ego}$ is the corresponding egocentric video of the exocentric video $\mathbf{x}_{exo}$,
and $\mathbf{K}_{exo}, [\mathbf{R}_{exo}, \mathbf{t}_{exo}]$ and $\mathbf{K}_{ego}, [\mathbf{R}_{ego}, \mathbf{t}_{ego}]$ are the camera (intrinsic and extrinsic) parameters of third and first views, respectively. Then, the procedure of obtaining the videos can be formed as a rendering function, as in Eqn. \eqref{eqn:render}.
\begin{equation} \label{eqn:render}
\small
\begin{split}
    \mathbf{x}_{exo} = \mathcal{R}(\mathbf{K}_{exo}, [\mathbf{R}_{exo},\mathbf{t}_{exo}])       \\ 
    \mathbf{\bar{x}}_{ego} = \mathcal{R}(\mathbf{K}_{ego}, [\mathbf{R}_{ego},\mathbf{t}_{ego}])
\end{split}
\end{equation}
where $\mathcal{R}$ is a rendering function that obtains the video with the given corresponding camera matrix and position. 
In Eqn. \eqref{eqn:render}, the rendering function $\mathcal{R}$ remains the same across views as $\mathbf{x}_{exo}$ and $\mathbf{\bar{x}}_{ego}$
are the pair video of the same scene. Moreover, as matrices represent the camera parameters, there exist linear transformations of the cameras between two views defined as in Eqn. \eqref{eqn:cam_equi}.
\begin{equation} \label{eqn:cam_equi}
\small
    \begin{split}
        \mathbf{K}_{ego} &= \mathbf{T}_{\mathbf{K}} \times \mathbf{K}_{exo} \\
        [\mathbf{R}_{ego}, \mathbf{t}_{ego}] &= \mathbf{T}_{\mathbf{Rt}} \times [\mathbf{R}_{exo}, \mathbf{t}_{exo}]
    \end{split}
\end{equation}

\noindent
\textbf{Remark 1: \textit{Cross-view Geometric Transformation}}
\textit{From Eqn. \eqref{eqn:render} and Eqn. \eqref{eqn:cam_equi}, we have observed that there exists a geometric transformation $\mathcal{T}$ of videos (images) between two camera views as follows:}
\begin{equation}
\small
    \mathbf{\bar{x}}_{ego} = \mathcal{T}(\mathbf{x}_{exo}; \mathbf{T_K}, \mathbf{T_{Rt}})
\end{equation}

In our proposed method, we consider the action recognition backbone model $F$ designed as a Transformer with self-attention layers. 
Given a video, the input of the Transformer is represented by $N+1$ tokens, including $N = \frac{T}{K}\frac{H}{P}\frac{W}{P}$ non-overlapped patches ($K \times P \times P$ is the patch size of the token) of a video and a single classification token. 
Let $\mathbf{a}_{exo}, \mathbf{\bar{a}}_{ego} \in \mathbb{R}^{\frac{T}{K} \times \frac{H}{P} \times \frac{W}{P}}$ be an attention map of the video frames w.r.t the classification token extracted from the network $F$ on the inputs $\mathbf{x}_{exo}$ and $\mathbf{\bar{x}}_{exo}$, respectively. The attention maps $\mathbf{a}_{exo}$ and $\mathbf{\bar{a}}_{ego}$ represent the focus of the model on the video over time w.r.t to the model predictions.
It should be noted the video and its attention map could be considered as a pixel-wised correspondence. Even though the patch size is greater than 1 ($K, P > 1$), a single value in the attention map always corresponds to a group of pixels in its patch. 
Therefore, without a lack of generality, with the changes of cameras from the exocentric view to the eccentric view, we argue that the focuses of the model (the attention maps) also change correspondingly to the transitions of the videos across views because both videos are representing the same action scene from different camera views. As a result, the transformation between two attention maps, i.e., $\mathbf{a}_{exo}$ and $\mathbf{\bar{a}}_{ego}$, can also be represented by a transformation 
$\mathcal{T}'$ w.r.t. the camera transformation matrices $\mathbf{T_K}$ and $\mathbf{T_{Rt}}$.

\noindent
\textbf{Remark 2: \textit{Cross-view Equivalent Transformation of Videos and Attentions}} 
{We argue that the transformations $\mathcal{T}$ and $\mathcal{T'}$ remain similar ($\mathcal{T} \equiv \mathcal{T'}$) as they are both the transformation interpolation based on the camera transformation matrices $\mathbf{T_K}$ and $\mathbf{T_{Rt}}$. Hence, 
the transformation $\mathcal{T}$ could be theoretically adopted to the attention transformation.}
\begin{equation}
\small
    \begin{split}
        \mathbf{\bar{a}}_{ego} = \mathcal{T}'(\mathbf{a}_{exo}; \mathbf{T_K}, \mathbf{T_{Rt}}) \equiv \mathcal{T}(\mathbf{a}_{exo}; \mathbf{T_K}, \mathbf{T_{Rt}})
    \end{split}
\end{equation}

Following the above remarks, we further consider the cross-view correlation between the videos and the attention maps.
Let $\mathcal{D}_x(\mathbf{x}_{exo}, \mathbf{\bar{x}}_{ego})$ and $\mathcal{D}_a(\mathbf{a}_{exo}, \mathbf{\bar{a}}_{ego})$ be the metrics measure the cross-view correlation in videos ($\mathbf{x}_{exo}, \mathbf{\bar{x}}_{ego}$) and attention maps ($\mathbf{a}_{exo}, \mathbf{\bar{a}}_{ego}$), respectively. 

From Remark 1 and Remark 2, we have observed that the transformation of both video and attention from the exocentric view to the egocentric view is represented by the shared transformation $\mathcal{T}$ and the camera transformation matrices $\mathbf{T_K}, \mathbf{T_{Rt}}$. 
In other words, the cross-view relation between  $\mathcal{D}_x(\mathbf{x}_{exo}, \mathbf{\bar{x}}_{ego})$ and $\mathcal{D}_a(\mathbf{a}_{exo}, \mathbf{\bar{a}}_{ego})$ relies on the shared transformation $\mathcal{T}(\cdot, \mathbf{T_K}, \mathbf{T_{Rt}}$) and the difference between $\mathbf{x}_{exo}$ and $\mathbf{a}_{exo}$. 
Therefore, we argue that the cross-view video correlation $\mathcal{D}_x(\mathbf{x}_{exo}, \mathbf{\bar{x}}_{ego})$  is theoretically proportional to the cross-view attention correlation $\mathcal{D}_a(\mathbf{a}_{exo}, \mathbf{\bar{a}}_{ego})$. In addition, the transformations between the two cameras are linear, as indicated in Eqn. \eqref{eqn:cam_equi}.
Thus, in our work, the proportion between $\mathcal{D}_x(\mathbf{x}_{exo}, \mathbf{\bar{x}}_{ego})$ and $\mathcal{D}_a(\mathbf{a}_{exo}, \mathbf{\bar{a}}_{ego})$ can be theorized as a linear relation and modeled by a linear scale $\alpha$ as in Eqn. \eqref{eqn:cross_view_condition}.
\begin{equation} \label{eqn:cross_view_condition}
\small
\begin{split}
    \mathcal{D}_x(\mathbf{x}_{exo}, \mathbf{\bar{x}}_{ego}) &\propto \mathcal{D}_a(\mathbf{a}_{exo}, \mathbf{\bar{a}}_{ego})\\
    \Leftrightarrow
    \mathcal{D}_x(\mathbf{x}_{exo}, \mathbf{\bar{x}}_{ego})  &= \alpha \mathcal{D}_a(\mathbf{a}_{exo}, \mathbf{\bar{a}}_{ego})
\end{split}
\end{equation}

\subsection{Unpaired Cross-View Self-Attention Loss}

\begin{figure*}
    \centering
    \includegraphics[width=1.0\textwidth]{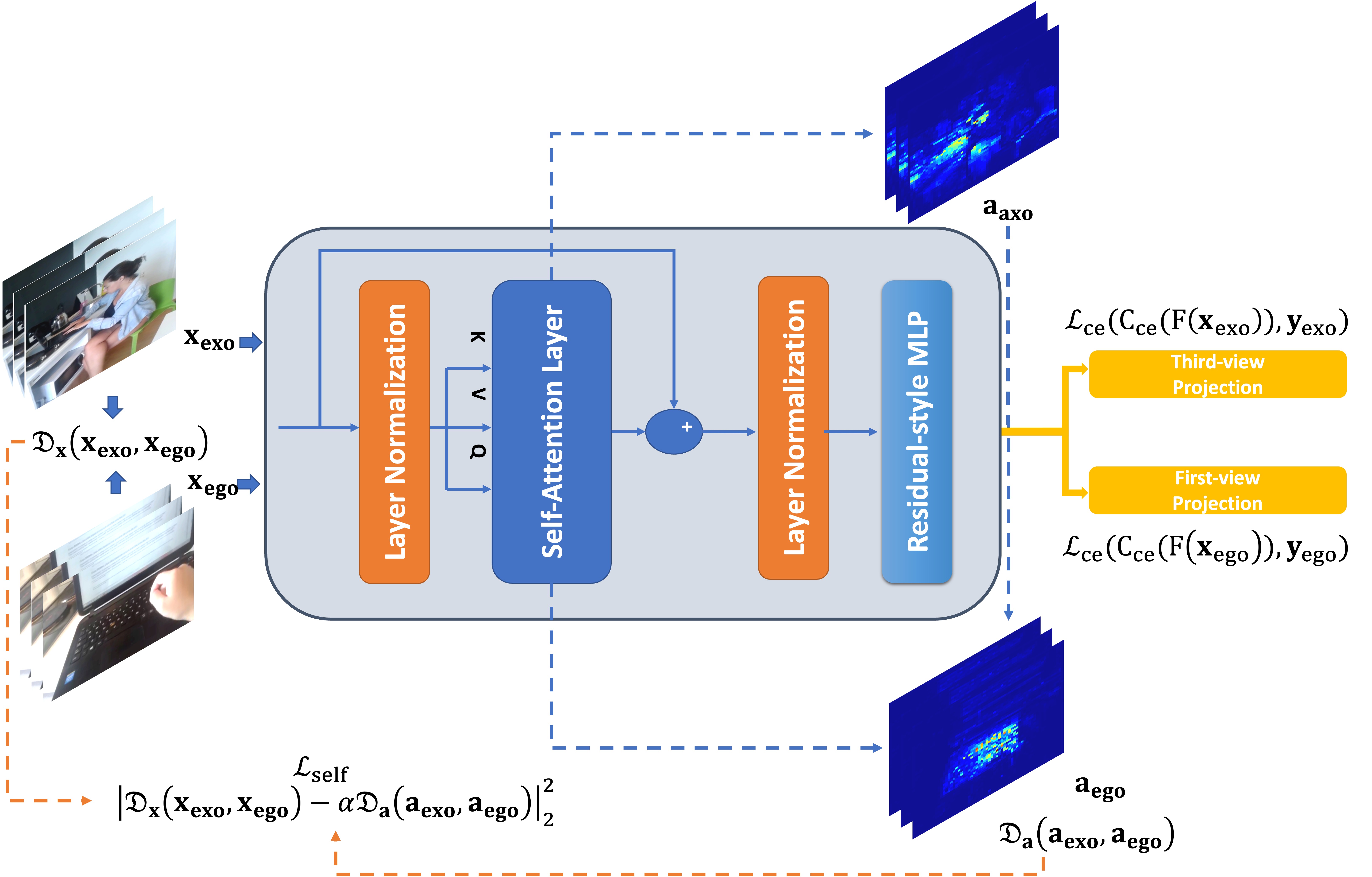}
    \caption{\textbf{The Proposed Framework.} The input videos $\mathbf{x}_{exo}$ and $\mathbf{x}_{ego}$ are first forwarded to Transformer $F$ followed by the corresponding classifiers $C_{exo}$ and $C_{ego}$, respectively. Then, the supervised cross-entropy loss $\mathcal{L}_{ce}$ is applied to the predictions produced by the model.
    Meanwhile, the attention maps of video inputs, i.e., $\mathbf{a}_{exo}$ and $\mathbf{a}_{ego}$,  are extracted and imposed by the cross-view self-attention loss $\mathcal{L}_{self}$. 
    }
    \label{fig:framework}
\end{figure*}

Eqn. \eqref{eqn:cross_view_condition} defines a condition that explicitly models the self-attention correlation based on the geometric transformation across views. Thus, to efficiently learn the action recognition model from the exocentric to the egocentric view, Eqn \eqref{eqn:general_opt} can be optimized w.r.t the condition in Eqn. \eqref{eqn:cross_view_condition}  and presented as in Eqn. \eqref{eqn:general_opt_with_cond}.
\begin{equation} \label{eqn:general_opt_with_cond}
\small
\begin{split}
    \arg\min_{\theta_F, \theta_{C_{exo}}, \theta_{C_{ego}}}[\mathbb{E}_{\mathbf{x}_{exo}, \mathbf{y}_{ego}}\mathcal{L}_{ce}(C_{exo}(F(\mathbf{x}_{exo})), \mathbf{y}_{exo}) \\ 
    + \mathbb{E}_{\mathbf{x}_{ego}, \mathbf{y}_{ego}}\mathcal{L}_{ce}(C_{ego}(F(\mathbf{x}_{ego})), \mathbf{y}_{ego})] \\
    s.t. \quad\quad \mathcal{D}_x(\mathbf{x}_{exo}, \mathbf{\bar{x}}_{ego})  = \alpha \mathcal{D}_a(\mathbf{a}_{exo}, \mathbf{\bar{a}}_{ego})
\end{split}
\end{equation}
Hence, optimizing Eqn. \eqref{eqn:general_opt_with_cond} can be solved by considering the cross-view constraint as a regularizer during training, i.e., $|| \mathcal{D}_x(\mathbf{x}_{exo}, \mathbf{\bar{x}}_{ego})  - \alpha \mathcal{D}_a(\mathbf{a}_{exo}, \mathbf{\bar{a}}_{ego})||_2^2$. However, it is noted that training requires a pair of third-view and first-view videos. Meanwhile, in practice, the video data of these two views are often recorded independently. Thus, optimizing Eqn. \eqref{eqn:general_opt_with_cond} by imposing the constraint of Eqn. \eqref{eqn:cross_view_condition} on pair data remains an ill-posed problem.
Instead of solving Eqn. \eqref{eqn:general_opt_with_cond} on pair data, let us consider all cross-view unpaired samples $(\mathbf{x}_{exo}, \mathbf{x}_{ego})$.
In addition, we assume that the cross-view correlation of videos $\mathcal{D}_x$ and attention maps $\mathcal{D}_{a}$ is bounded by a certain threshold $\beta$, i.e.,  $\forall \mathbf{x}_{exp},  \mathbf{x}_{ego}: \mathcal{D}_x(\mathbf{x}_{exo}, \mathbf{x}_{ego}) \leq \beta$ and $\forall \mathbf{a}_{exp},  \mathbf{a}_{ego}: \mathcal{D}_a(\mathbf{a}_{exo}, \mathbf{a}_{ego}) \leq \beta$.
This assumption implies that the distribution shifts (i.e., the changes of views) from the exocentric to the egocentric view are bounded to ensure that the model can generalize its capability across views. Hence, our \textbf{\textit{Cross-view Self-Attention Loss}} on unpaired data can be formulated as in Eqn. \eqref{eqn:loss_for_unpair}.
\begin{equation}\label{eqn:loss_for_unpair} 
\small
\begin{split}
    \mathcal{L}_{self} = \mathbb{E}_{\mathbf{x}_{exo}, \mathbf{x}_{ego}} \lambda||\mathcal{D}_{x}(\mathbf{x}_{exo}, \mathbf{x}_{ego}) - 
    \alpha\mathcal{D}_{a}(\mathbf{a}_{exo}, \mathbf{a}_{ego})||_2^2
\end{split}
\end{equation}
where $\lambda$ is the hyper-parameter controlling the relative importance of $\mathcal{L}_{self}$.
Intuitively, even though the pair samples between exocentric and egocentric views are inaccessible, the cross-view constraints between videos and attention maps can still be imposed by modeling the topological constraint among unpaired samples. Furthermore, under our cross-view distribution shift assumption, our loss in Eqn. \eqref{eqn:loss_for_unpair} can be proved as an upper bound of the constrain Eqn. \eqref{eqn:cross_view_condition} on pair samples as follows:
\begin{equation} \label{eqn:upper_bound}
\small
\begin{split}
    \mathcal{D}_{x}(\mathbf{x}_{exo}, &\mathbf{\bar{x}}_{ego}) - \alpha\mathcal{D}_{a}(\mathbf{a}_{exo}, \mathbf{\bar{a}}_{ego}) 
     \leq \mathcal{D}_{x}(\mathbf{x}_{exo}, \mathbf{x}_{ego})    -\alpha\mathcal{D}_{a}(\mathbf{a}_{exo}, \mathbf{a}_{ego})  + (1+\alpha)\beta
\end{split}
\end{equation}
Eqn. \eqref{eqn:upper_bound} can be proved using the triangle inequality property of $\mathcal{D}_{x}$ and $\mathcal{D}_{a}$. 
Eqn. \eqref{eqn:upper_bound} can be proven as follows. 
Since $\mathcal{D}_{x}$ and $\mathcal{D}_{a}$ are the metrics that measure the correlation of videos and attention maps, respectively; therefore, for all $\mathbf{x}_{ego}$ and $\mathbf{a}_{ego}$, these metrics have to satisfy the triangular inequality as follows:
\begin{equation} \label{eqn:triangluar}
\begin{split}
    \mathcal{D}_{x}(\mathbf{x}_{exo}, \mathbf{x}_{ego}) + \mathcal{D}_{x}(\mathbf{x}_{ego}, \mathbf{\bar{x}}_{ego})  &\geq \mathcal{D}_{x}(\mathbf{x}_{exo}, \mathbf{\bar{x}}_{ego}) \\
    \mathcal{D}_{a}(\mathbf{a}_{exo}, \mathbf{a}_{ego}) + \mathcal{D}_{a}(\mathbf{a}_{ego}, \mathbf{\bar{a}}_{ego})  &\geq \mathcal{D}_{a}(\mathbf{a}_{exo}, \mathbf{\bar{a}}_{ego}) \\
\end{split}
\end{equation}
In addition, under our  cross-view distribution shift assumption, the metrics $\mathcal{D}_\mathbf{x}$ and $\mathcal{D}_\mathbf{y}$ are bounded by a threshold $\beta$, i.e., $ \mathcal{D}_{x}(\mathbf{x}_{exo}, \mathbf{x}_{ego}) \leq \beta$ and $\mathcal{D}_{a}(\mathbf{a}_{exo}, \mathbf{a}_{ego}) 
 \leq \beta$. As a result, the cross-view self-attention constraint can be further extended as follows:
\begin{equation} \label{eqn:upper_bound_2}
\begin{split}
    &\mathcal{D}_{x}(\mathbf{x}_{exo}, \mathbf{\bar{x}}_{ego}) - \alpha\mathcal{D}_{a}(\mathbf{a}_{exo}, \mathbf{\bar{a}}_{ego}) \\
    &\leq \mathcal{D}_{x}(\mathbf{x}_{exo}, \mathbf{x}_{ego}) + \mathcal{D}_{x}(\mathbf{x}_{ego}, \mathbf{\bar{x}}_{ego}) - \alpha\mathcal{D}_{a}(\mathbf{a}_{exo}, \mathbf{\bar{a}}_{ego})\\
    &\leq \mathcal{D}_{x}(\mathbf{x}_{exo}, \mathbf{x}_{ego}) + \beta - \alpha(\mathcal{D}_{a}(\mathbf{a}_{exo}, \mathbf{a}_{ego})+\beta)+\alpha\beta \\
    &\leq \mathcal{D}_{x}(\mathbf{x}_{exo}, \mathbf{x}_{ego})   -\alpha\mathcal{D}_{a}(\mathbf{a}_{exo}, \mathbf{a}_{ego}) + (1+\alpha)\beta
\end{split}
\end{equation}

As shown in Eqn. \eqref{eqn:upper_bound}, as $\mathcal{D}_{x}(\mathbf{x}_{exo}, \mathbf{x}_{ego})   -\alpha\mathcal{D}_{a}(\mathbf{a}_{exo}, \mathbf{a}_{ego}) + (1+\alpha)\beta$ is the upper bound of $\mathcal{D}_{x}(\mathbf{x}_{exo}, \mathbf{\bar{x}}_{ego}) - \alpha\mathcal{D}_{a}(\mathbf{a}_{exo}, \mathbf{\bar{a}}_{ego})$, minimizing $||\mathcal{D}_{x}(\mathbf{x}_{exo}, \mathbf{x}_{ego})  - \alpha\mathcal{D}_{a}(\mathbf{a}_{exo}, \mathbf{a}_{ego})||_2^2$ also imposes the constraint of $||\mathcal{D}_{x}(\mathbf{x}_{exo}, \mathbf{\bar{x}}_{ego}) - \alpha\mathcal{D}_{a}(\mathbf{a}_{exo}, \mathbf{\bar{a}}_{ego})||_2^2$. Note that $\alpha$ and  $\beta$ are constant numbers, which can be excluded during training.
Therefore, the constraints of cross-view correlation on pair samples in  Eqn. \eqref{eqn:general_opt_with_cond} is guaranteed when optimizing $\mathcal{L}_{self}$ defined in Eqn. \eqref{eqn:loss_for_unpair}. More importantly, our proposed cross-view self-attention loss\textbf{\textit{ does NOT require the pair data between exocentric and egocentric views}} during training.
Fig. \ref{fig:framework} illustrates our proposed cross-view learning framework.

\noindent
\textbf{Cross-view Topological Preserving Property:} The proposed loss defined in Eqn. \eqref{eqn:loss_for_unpair} to impose the cross-view correlation over all unpaired samples is a special case of the Gromov-Wasserstein \cite{GW_distance} distance between the video and the attention map distributions where the association matrix has been pre-defined. As a result, our loss inherits these Gromov-Wasserstein properties to preserve the topological distributions between the video and attention space.
Remarkably, the cross-view topological structures of video distributions are preserved in cross-view attention distributions.

\subsection{The Choices of Correlation Metrics}

As shown in Eqn. \eqref{eqn:loss_for_unpair}, the choice of correlation metric $\mathcal{D}_{x}$ and $\mathcal{D}_{a}$ is one of the primary factors directly influencing the performance of the action recognition models. The direct metrics, i.e., $\ell_2$, could be straightforwardly adopted for the correlation metric $\mathcal{D}_x$ and $\mathcal{D}_a$. However, this direct approach is ineffective because the deep semantic information of videos is not well modeled in the direct Euclidean metric $\ell_s$. To overcome this limitation, we propose designing $\mathcal{D}_x$ as the correlation metric on the deep latent spaces defined as in Eqn. \eqref{eqn:dx_G}.
\begin{equation}
\label{eqn:dx_G}
\small
\mathcal{D}_{x}(\mathbf{x}_{exo}, \mathbf{x}_{ego}) = \mathcal{D}^G_{x}(\mathbf{x}_{exo}, \mathbf{x}_{ego}) = || G(\mathbf{x}_{exo}) - G(\mathbf{x}_{ego}) ||^2_2
\end{equation}
where $G: \mathbb{R}^{T \times H \times W \times 3} \to \mathbb{R}^{K}$ be the deep network trained on the large-scale dataset. %
Intuitively, measuring the correlation between two videos provides a higher level of semantic information since the deep representation extracted by the large pre-trained model $G$ captures more contextual information about the videos \cite{johnson2016perceptual, duong2020vec2face}.

As $\mathcal{D}_a$ measures the correlation between two attention maps where, each of which is in the form of the probability distribution, $\mathcal{D}_a$ should be defined as the statistical distance to measure the correlation between two probabilistic attention maps comprehensively. Thus, we propose designing $\mathcal{D}_{a}$ as the Jensen-Shannon divergence defined in Eqn. \eqref{eqn:da_JS}.
\begin{equation} \label{eqn:da_JS}
\small
\begin{split}
    \mathcal{D}_{a}(\mathbf{a}_{exo}, \mathbf{a}_{ego}) &= \mathcal{D}^{JS}_a(\mathbf{a}_{exo}, \mathbf{a}_{ego}) \\
    &= \frac{1}{2}(\mathcal{D}_{KL}(\mathbf{a}_{exo} || \mathbf{a}_{ego}) + \mathcal{D}_{KL}(\mathbf{a}_{ego} || \mathbf{a}_{exo})) 
\end{split}
\end{equation}
where $\mathcal{D}_{KL}$ is the  Kullback–Leibler divergence.
To satisfy the cross-view distribution shift assumption aforementioned, 
the correlation metrics $\mathcal{D}_{x}$ and $\mathcal{D}_{a}$ are constrained by the threshold $\beta$, i.e., $\mathcal{D}_{x}(\mathbf{x}_{exo}, \mathbf{x}_{ego}) = \min\left(\mathcal{D}^G_{x}(\mathbf{x}_{exo}, \mathbf{x}_{ego}), \beta\right)$ and $\mathcal{D}_{a}(\mathbf{a}_{exo}, \mathbf{a}_{ego}) = \min\left(\mathcal{D}^{JS}_{a}(\mathbf{a}_{exo}, \mathbf{a}_{ego}), \beta\right)$. In our experiments, the value of $\beta$ is set to $200$.

\section{Experimental Results}

This section first briefly presents the datasets and the implementation details in our experiments. Then, we analyze the effectiveness of the approach in ablative experiments, followed by comparing results with prior methods on the standard benchmarks of first-view action recognition.

\subsection{Datasets and Implementation Details}

Following the common practice in action recognition \cite{swin, slowfast, timesformer}, Kinetics has been used as the third-view dataset in our experiment due to its large scale and diverse actions. 
To evaluate the effectiveness of our approach, we use EPIC-Kitchens and Charades-Ego as our first-view datasets. These two datasets are currently known as large-scale and challenging benchmarks in egocentric action recognition.

\noindent
\textbf{Kinetics-400} \cite{kinetics} is a large-scale third-view action recognition dataset including 300K videos of 400 classes of human actions.
The dataset is licensed by Google Inc. under a Creative Commons Attribution 4.0 International License.

\noindent
\textbf{Charades-Ego} \cite{charades-ego} is a first-view action recognition dataset that consists of 157 action classes with 68K clips. The license of Charades-Ego is registered for academic purposes by the Allen Institute for Artificial Intelligence.

\noindent
\textbf{EPIC-Kitchens-55} \cite{epic} is a large-scale multi-task egocentric dataset of daily activities in kitchens. 
The action recognition task includes 55 hours of 39K clips and is annotated by interactions between 352 nouns and 125 verbs.

\noindent
\textbf{EPIC-Kitchens-100}~\cite{epic-100} is an larger version of the EPIC-Kitchens-55 where it is extended to 100 hours of 90k action clips. Each single action segment is annotated by an action of 97 verbs and 300 nouns. The EPIC Kitchens dataset was published under the Creative Commons Attribution-NonCommerial 4.0 International License.

\noindent
\textbf{NTU RGB+D}~\cite{shahroudy2016ntu} is the RGB-D human action recognition dataset. The dataset consists of $56,880$ samples of $60$ action classes collected from $40$ subjects. Each action is captured using three cameras with different angles, i.e., $-45^o$, $0^o$, and $+45^o$.

\noindent
\textbf{Evaluation Metrics} Our experiments follow the standard benchmarks of the Charades-Ego and EPIC-Kitchens for action recognition. We report the mean average precision (mAP) in the Charades-Ego \cite{charades-ego} experiments and Top 1 and Top 5 accuracy of verb, noun, and action predictions of the validation set in EPIC-Kitchens \cite{epic-100, epic} experiments.

\begin{table}[!b]
\caption{\textbf{Effectiveness of the Scale $\alpha$ in the Linear Relation to the Charades-Ego (E-Ego) and EPIC-Kitchen-55 (EPIC) Action Recognition Benchmarks.}}
\label{tab:alpha_ab}

\centering
\begin{tabular}{|c|c|c|c|c|c|}
\hline
\multirow{2}{*}{$\alpha$}      &       C-Ego       & \multicolumn{2}{c|}{EPIC Verb}         & \multicolumn{2}{c|}{EPIC Noun}         \\
\cline{2-6}
 & mAP & Top  1 & Top 5 & Top 1 & Top 5 \\
 \hline
0.00  & 20.70        & 41.94  & 67.31 &	43.19 &	60.14 \\
0.25  & 25.09        & 55.96  & 89.37 &	55.96 & 80.65 \\
0.50  & 28.97        & 58.84  & 87.24 & 54.75 & 75.27 \\
0.75  & \textbf{31.95}        & 60.80  & 89.62 & 57.42 & 77.77 \\
1.00  & 30.68        & 68.97  & 89.53 & 44.87 & 70.98 \\
1.50  & 29.51        & \textbf{73.52}	& \textbf{92.22}	& \textbf{68.19}	& \textbf{84.93} \\ 
2.00  & 27.80     & 69.60	& 92.54	& 61.60	& 81.22 \\
\hline
\end{tabular}
\end{table}

\noindent
\textbf{Implementation}
In our work, we adopt the design of the Vision Transformation Base model (ViT-B) \cite{vit} for our Transformer backbone. Our model is implemented in Python using the PyTorch and PySlowFast \cite{pyslowfast} frameworks. The input video of our network consists of $T = 16$ frames sampled at the frame rate of $1/4$, and the input resolution of each video frame is $H\times W = 224\times 224$. Each video is tokenized by the non-overlapping patch size of $K \times P \times P = 2 \times 16 \times 16$. Each token is projected by an embedding where the dimension length of the embedding is set to $768$. 
Our model has $12$ Transformer layers, and the number of heads in each self-attention layer is set to $8$. 
The Stochastic Gradient Descent algorithm optimizes the entire framework, where our models are trained for $50$ epochs. The cosine learning policy is utilized in our training, where the base learning rate is set to $0.00125$. 
Similar to \cite{MViTv2, slowfast}, we also apply several augmentation methods during training to increase the diversity of training data. All of our models are trained on the four 40GB-VRAM A100 GPUs, and the batch size in each GPU is set to $4$.
Swin-B \cite{swin} pre-trained on the Kinetics-400 dataset has been adopted for our network $G$ in Eqn. \eqref{eqn:dx_G}. Since we do not want the gradients produced by the supervised loss $\mathcal{L}_{ce}$ being suppressed by the cross-view loss $\mathcal{L}_{self}$, the hyper-parameter $\lambda$ is set to $5.10^{-3}$.
In our evaluation, following prior works
\cite{ego-exo, timesformer}, each input is sampled in the middle of the video. The final result from the video input is obtained by averaging the prediction scores of three spatial crops, i.e.,  top-left, center, and bottom-right.

\subsection{Ablation Studies}

Our ablative experiments report the results of our CVAR method with different settings trained on the Kinetics-400 $\to$ Charades-Ego and Kinetics-400 $\to$ EPIC-Kitchens-55 benchmarks. All the models are trained with the same learning configuration for fair comparisons. %

\begin{table}[!b]
\centering
\caption{\textbf{Effectiveness of the Choices of Correlation Metrics to the Charades-Ego (E-Ego) and EPIC-Kitchen-55 (EPIC) Action Recognition Benchmarks.}}
\label{tab:metric_ab}
\setlength{\tabcolsep}{3pt}
\def\arraystretch{1.1}
\begin{tabular}{|cc|cc|c|cc|cc|}
\hline

\multicolumn{2}{|c|}{$\mathcal{D}_{x}$}                       & \multicolumn{2}{c|}{$\mathcal{D}_{a}$}                       & C-Ego &  \multicolumn{2}{|c}{EPIC Verb}       & \multicolumn{2}{|c|}{EPIC Noun}      \\
\cline{1-4}\cline{5-9}
$\ell_2$                    & $\mathcal{D}_x^{G}$                   & $\ell_2$                    & $\mathcal{D}_a^{JS}$                   & mAP & Top  1 & Top 5 & Top 1 & Top 5 \\
\hline
\cmark &                       & \cmark &                       & 27.80        & 60.97  & 89.95 & 58.05 & 78.07 \\
\cmark &                       &                       & \cmark & 28.77        & 61.13  & 90.16 & 58.05 & 78.40 \\
                      & \cmark & \cmark &                       & 29.11  & 63.13	& 90.12	& 59.68	& 80.03 \\
                      & \cmark &                       & \cmark & \textbf{31.95}        & \textbf{73.52}  & \textbf{92.22} & \textbf{68.19}	& \textbf{84.93} \\
\hline
\end{tabular}
\end{table}

\noindent
\textbf{Effectiveness of the scale $\alpha$} 
In this experiment, the metrics defined in Eqn. \eqref{eqn:dx_G} and Eqn. \eqref{eqn:da_JS} have been adopted to $\mathcal{D}_{x}$ and $\mathcal{D}_{a}$. The value $\alpha$ ranges from 0.0 to 2.0. When $\alpha = 0.0$, it is equivalent to ViT simultaneously trained on both third-view and first-view datasets.
As shown in Table \ref{tab:alpha_ab}, the mAP performance on the Charades-Ego benchmark is consistently improved when the value of $\alpha$ increases from $0.1$ to $0.75$ and achieves the best performance at the value of $\alpha = 0.75$ and the mAP performance is $31.95\%$. 
Similarly, on the EPIC-Kitchen-55 benchmarks, the Top 1 and Top 5 accuracy is gradually improved w.r.t the increasing of $\alpha$ and reaches the maximum performance when the value of $\alpha$ is $1.50$ in which the Top 1 accuracy on EPIC Verb and EPIC Noun are $73.52\%$ and $68.19\%$. 
Then, the performance on both benchmarks steadily decreases when the value of $\alpha$ keeps increasing over the optimal point. 
Indeed, the variation in the video space is typically higher than in the attention maps due to the higher complexity of video data where the video data contains much more information, e.g., objects, humans, and interactions, etc.; meanwhile, the attention maps represent the focus of the models w.r.t model decisions. Thus, if the value of $\alpha$ is small, it could not represent the correct proportion of changes between videos and attention maps.
Meanwhile, the higher value of $\alpha$ inclines to exaggerate the model focuses, i.e., attention maps, that results in the performance.

\noindent
\textbf{Effectiveness of the metrics} 
This experiment studies the effectiveness of correlation metrics on the performance of the action recognition models on first-view videos.
The optimal value of the linear $\alpha$ in the previous ablation study has been adopted in this experiment.
For each metric correlation, we study its effect by comparing the performance of action recognition models using our metric in Eqn. \eqref{eqn:dx_G} and Eqn. \eqref{eqn:da_JS} against the Euclidean distance $\ell_2$.
As our results in Table \ref{tab:metric_ab}, by measuring the correlation of videos on the deep latent spaces, i.e., $\mathcal{D}_x^G$, the performance of the action recognition model has been improved, e.g., $28.77\%$ to $31.95\%$ (results using $\mathcal{D}_a^{JS}$). This improvement is gained thanks to the deep semantic representation extracted by deep network $G$.
Besides, the probability metric used to measure the correlation between attention maps, i.e., $\mathcal{D}_a^{JS}$, has illustrated its significant role. For example, the performance of the model has been promoted by $+2.84\%$ from $29.11\%$ (using $\ell_2$) to $31.95\%$ (using $\mathcal{D}_a^{JS}$). As the attention map is the probability distribution, using the Jensen-Shannon divergence as the correlation metric provides the informative difference of the model's focus over the videos. Meanwhile, $\ell_2$ tends to rely on the difference of the magnitude of the attention, which provides less correlation information between two attentions.
Figure \ref{fig:effectiveness_metric} below shows that attention has been effectively learned and transferred from the third to the first view.

\begin{table}[!b]
\centering
\caption{\textbf{Effectiveness of the Transformer Layers to the Charades-Ego (E-Ego) and EPIC-Kitchen-55 (EPIC) Action Recognition Benchmarks.}}
\label{tab:attention_ab}
\setlength{\tabcolsep}{2.5pt}
\begin{tabular}{|cccc|c|cc|cc|}
\hline

\multicolumn{4}{|c|}{Transformer Layers} & C-Ego &  \multicolumn{2}{|c}{EPIC Verb}       & \multicolumn{2}{|c|}{EPIC Noun}      \\
\cline{5-9}
 1-3 & 4-6 & 7-9 & 10-12 & mAP & Top  1 & Top 5 & Top 1 & Top 5 \\
\hline
\cmark & & & & 25.65 & 60.47 & 90.26 & 57.85 & 78.58 \\
\cmark & \cmark & & & 28.19 & 68.46 & 91.08 & 66.54 & 83.36 \\
\cmark & \cmark & \cmark & & 30.60 & 69.27 & \textbf{92.58} & 68.09 & \textbf{85.02} \\
\cmark & \cmark & \cmark & \cmark & \textbf{31.95}        & \textbf{73.52}  & 92.22 & \textbf{68.19}	& 84.93 \\
\hline
\end{tabular}
\end{table}

\begin{figure}[!t]
    \centering
    \includegraphics[width=1.0\textwidth]{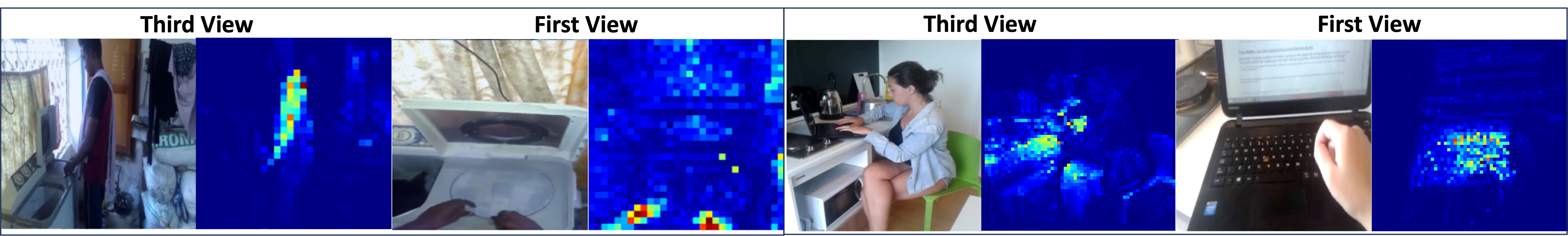}
    \caption{Effectiveness of Our Metrics in Cross-view Learning}
    \label{fig:effectiveness_metric}
\end{figure}

\begin{figure}[!t]
    \centering
    \includegraphics[width=1.0\textwidth]{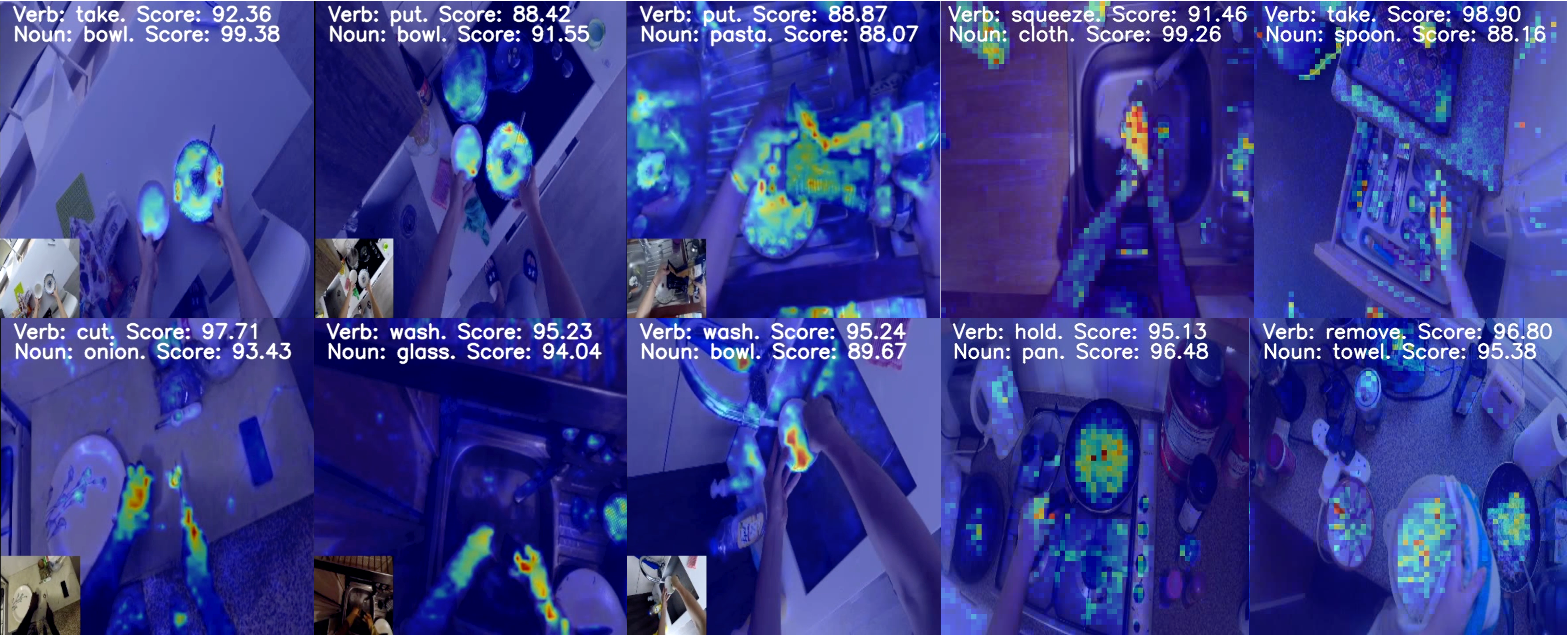}
    \caption{Attention Visualization of Model Prediction on EPIC Kitchen Videos.}
    \label{fig:att_vis}
\end{figure}

\noindent
\textbf{Effectiveness of Transformer Layers} This experiment studies the effectiveness of imposing the cross-view loss into attention maps of the Transformer layers. In this experiment, we adopt the optimal setting of the linear scale ($\alpha$) and correlation metrics ($\mathcal{D}_x$, $\mathcal{D}_a$) in the previous ablation studies.
We consider four groups of Transformer layers, each consisting of three consecutive layers, i.e., Layer 1-3, Layer 4-6, Layer 7-9, and Layer 10-12. As experimental results in Table \ref{tab:attention_ab}, the later Transformer layers of our model play an important role than the initial ones. In particular, when imposing the cross-view loss on only the first three Transformer layers, the performance of Charades-Ego has achieved 25.65\% and the Top 1 accuracy of verb and noun predictions in EPIC-Kitchens-55 is 60.47\% and 57.85\%. Meanwhile, enforcing the cross-view self-attention loss into all attention layers brings better performance and achieves the best performance, i.e., the mAP of 31.95\% on Charades-Ego and Top 1 accuracy of 73.52\% and 68.19\% on EPIC-Kitchens-55.
Fig. \ref{fig:att_vis} visualizes the attention maps of our model.

\begin{table}[!b]
\centering
\caption{Effectiveness of Different Networks}\label{tab:exp_net}
\begin{tabular}{|l|cc|cc|}
\hline
\multirow{2}{*}{Backbone}                 & \multicolumn{2}{c|}{EPIC-55 Verb} & \multicolumn{2}{c|}{EPIC-55 Noun} \\
\cline{2-5}
            & Top 1         & Top 5         & Top 1         & Top 5         \\
\hline
Swin-B \cite{swin} & 56.40 & 85.84 & 47.68 & 71.02 \\
\hline
ViT \cite{vit}               & 41.76         & 69.49         & 44.19         & 60.52         \\
Ensemble ViT + SwinB & 58.97	& 88.61 &	49.21	& 74.27\\
ViT+CVAR           & \textbf{73.52}         & 92.22         & \textbf{68.19}         & \textbf{84.93}         \\
\hline
TimeSFormer \cite{timesformer}        & 41.37         & 67.40         & 42.44         & 59.55         \\
Ensemble TimeSFormer + SwinB & 59.07 & 88.95	& 50.11 & 77.27 \\
TimeSFormer+CVAR & 72.17         & \textbf{95.19}         & 62.83         & 83.49        \\
\hline
\end{tabular}
\end{table}

\noindent
\textbf{Effectiveness of Different Network Backbone}
To further illustrate the robustness of CVAR against the network backbone and ensemble models, we further evaluate CVAR with different network backbones.
We report the experimental results of ViT and TimeSFormer \cite{timesformer} with and without our proposed geometric cross-view constraint.
We also report the result of the ensemble model with SwinB to illustrate the robustness of our approach compared to the ensemble approach.
The experimental results in Table \ref{tab:exp_net} have proved our proposed loss has robustly and consistently improved the performance of action recognition models. Moreover, our proposed CVAR significantly outperforms the ensemble model.
Our approach emphasizes a novel geometric cross-view correlation that can be adopted on top of other Transformers.

\noindent
{\textbf{Effectiveness of Paired and Unpaired Data.} 
To illustrate the effectiveness of our proposed approach on paired data, we conduct an ablation study on the Charades Ego dataset. The Charades Ego provides a pair of both exocentric and egocentric videos.
We evaluate our approach with two different settings of paired data, i.e., Charades-Exo $\to$ Charades-Ego, and unpaired data, i.e., Kinetics-400 $\to$ Charades-Ego.
We conduct our experiments using different metrics of $\mathcal{D}_x$ and $\mathcal{D}_a$.
As shown in Table \ref{tab:ablation_paired_unpaired}, the performance of the model using paired data is slightly better than the one using unpaired data.
However, since the data scale of Kinetics-400 is quite large compared to Charades-Exo, our approach using unpaired data achieves competitive performance.
The experimental results have shown that our proposed approach remains efficient in both cases of paired and unpaired data. In addition, if the source data is relatively large, the model's performance using unpaired data can even achieve competitive performance compared to the one using paired data.
}
    
\begin{table}[!t]
\centering
\caption{\textbf{Effectiveness of Paired and Unpaired Data to the Charades-Ego Action Recognition Benchmarks.}}
\label{tab:ablation_paired_unpaired}
\begin{tabular}{|cc|cc|cc|}
\hline

\multicolumn{2}{|c|}{$\mathcal{D}_{x}$}                       & \multicolumn{2}{c|}{$\mathcal{D}_{a}$}                       & Kinetics-400  &  Charades-Exo \\
$\ell_2$                    & $\mathcal{D}_x^{G}$                   & $\ell_2$                    & $\mathcal{D}_a^{JS}$                   & $\to$ Charades-Ego & $\to$ Charades-Ego \\
\hline
\cmark &                       & \cmark &                       & 27.80          &  28.48\\
\cmark &                       &                       & \cmark & 28.77          &  29.15 \\
                      & \cmark & \cmark &                       & 29.11          &  30.19 \\
                      & \cmark &                       & \cmark & 31.95 &  \textbf{32.64} \\
\hline
\end{tabular}
\end{table}

\subsection{Comparisons with State-of-the-Art Results}

\begin{wraptable}[14]{r}{0.4\textwidth}
\centering
\caption{\textbf{Comparisons on Charades-Ego.}}
\label{tab:charades_ego}
\resizebox{0.4\textwidth}{!}{
\begin{tabular}{|l|c|}
		\hline
		Method	& mAP	\\
		\hline
		ActorObserverNet~\cite{sigurdsson2018actor} & 20.00 \\
		SSDA~\cite{choi2020unsupervised} & 23.10 \\
        I3D~\cite{choi2020unsupervised} & 25.80 \\
        DANN \cite{ganin2016domain}  & 23.62 \\
        SlowFast~\cite{slowfast} & 25.93 \\
        Frozen \cite{frozen} & 28.80 \\
        MViT-V2 & 25.65 \\
        Swin-B \cite{swin} & 28.77 \\
        Ego-Exo  + ResNet-50 \cite{ego-exo} & 26.23 \\
        Ego-Exo  + SlowFast R50 \cite{ego-exo}  & 28.04 \\
        Ego-Exo*  + ResNet-50 \cite{ego-exo}  & 27.47 \\
        Ego-Exo* + SlowFast R50 \cite{ego-exo}  & 29.19 \\
        Ego-Exo* + SlowFast R101 \cite{ego-exo}  & 30.13 \\
        \textbf{CVAR (Ours)} & \textbf{ 31.95} \\
        \hline
\end{tabular}
}
\end{wraptable}
\noindent
\textbf{Kinetics-400 $\to$ Charades-Ego}
Table \ref{tab:charades_ego} presents results of our CVAR compared to prior methods, i.e., ActorObserverNet \cite{sigurdsson2018actor}, SSDA \cite{choi2020unsupervised}, I3D \cite{choi2020unsupervised}, DANN \cite{ganin2016domain}, SlowFast \cite{slowfast}, Frozen \cite{frozen}, MViT-V2 \cite{MViTv2}, Swin-B \cite{swin}, and Ego-Exo \cite{ego-exo}, on the Charades-Ego benchmark. 
Our results in Table \ref{tab:charades_ego} have gained SOTA performance where our mAP accuracy in our approach has achieved $31.95\%$.
Compared to direct training approaches \cite{vit, frozen, swin, mvit, choi2020unsupervised}, our method achieves better performance than other methods by a large margin, e.g., higher than Swin-B \cite{swin} by $3.18\%$.
Compared with the prior pre-training approach using additional egocentric tasks, our result is higher than Ego-Exo \cite{ego-exo} by $+1.82\%$.
Meanwhile, compared with domain adaptation approaches \cite{ganin2016domain, choi2020unsupervised}, our methods outperform DANN by $+8.33\%$. 

\begin{table}[!b]
\centering
\caption{\textbf{Comparisons on EPIC-Kitchen-55.}}
\label{tab:epic_55}
\label{tb:different_baseline}
\setlength{\tabcolsep}{3.5pt}
	\begin{tabular}{|l|cc|cc|}
		\hline
		\multirow{2}{*}{Method}   & \multicolumn{2}{c|}{EPIC verbs} & \multicolumn{2}{c|}{EPIC nouns} \\
		\cline{2-5}
		  & Top 1 & Top 5 & Top 1 & Top 5	\\
		\hline
        ResNet-50 \cite{slowfast}  & 61.19 & 87.49 & 46.18 & 69.72 \\
        MViT-V2 \cite{MViTv2} & 55.17	& 89.87	& 56.59	& 79.40 \\
        Swin-B \cite{swin} & 56.40 & 85.84 & 47.68 & 71.02 \\
        DANN \cite{ganin2016domain}  & 61.27 & 87.49 & 45.93 & 68.73 	  \\
        Joint-Embed \cite{sigurdsson2018actor} & 61.26 & 87.17 & 46.55 & 68.97	  \\
        Ego-Exo + ResNet-50 \cite{ego-exo}      & {62.83} & {87.63} & {48.15} & {70.28} \\
        Ego-Exo + SlowFast \cite{ego-exo} & 65.97	& 88.91	& 49.42	& 72.35 \\
        Ego-Exo* + ResNet-50 \cite{ego-exo}     & 64.26 & 88.45 & 48.39 & 70.68 \\
        Ego-Exo* + SlowFast \cite{ego-exo} & 66.43	& 89.16	& 49.79	& 71.60 \\
        \textbf{CVAR (Ours)} & \textbf{73.52}	& \textbf{92.22}	& \textbf{68.19}	& \textbf{84.93} \\
		\hline				
	\end{tabular}
\end{table}

\noindent
\textbf{Kinetics-400 $\to$ EPIC-Kitchens-55}
Table \ref{tab:epic_55} presents the results of our approach compared to prior methods, i.e., ResNet-50 \cite{slowfast}, DANN \cite{ganin2016domain}, SlowFast \cite{slowfast}, MViT-V2 \cite{MViTv2}, Swin-B \cite{swin}, and Ego-Exo \cite{ego-exo}, on the EPIC-Kitchens-55 benchmark. 
Our proposed CVAR has gained the SOTA performance where our Top 1 accuracy on EPIC Verb and EPIC Noun of our approach has achieved $73.52\%$ and $68.19\%$, respectively.
Our proposed approach outperforms the traditional direct training approaches \cite{slowfast, MViTv2, swin} by a large margin.
In addition, our result is higher than the pre-training approach using additional egocentric tasks, i.e., Ego-Exo \cite{ego-exo}, by $+6.48\%$ and $+18.4\%$ on Top 1 accuracy of verb and noun predictions. Our method also outperforms the domain adaptation approach \cite{ganin2016domain}.

\noindent
\textbf{Kinetics-400 $\to$ EPIC-Kitchens-100}
Table \ref{tab:epic_100} compares our results with TSN \cite{tsn}, TRN \cite{trn}, TBN \cite{epic-fusion}, TSM \cite{lin2019tsm}, SlowFast \cite{slowfast}, MViT-V2 \cite{MViTv2}, Ego-Exo using SlowFast-R50 \cite{ego-exo}, and Swin-B \cite{swin} on the EPIC-Kitchens-100 benchmark. 
Overall, our proposed CVAR has achieved the SOTA performance where the Top 1 accuracy of verb, noun, and action predictions are $69.37\%$, $61.03\%$, and $46.15\%$, respectively. Also, CVAR has gained competitive performance on the sets of unseen participants and tail classes.
Compared to prior direct training methods \cite{swin, MViTv2, vit}, out method outperforms these approaches by a notable margin, i.e., higher than Swin-B by $+1.44\%$ and $+2.34\%$ on Top 1 Accuracy of Verb and Noun predictions in overall.
Also, our results outperform Ego-Exo in overall accuracy and unseen participants and tail classes.

\begin{table}[!t]
    \centering
    \footnotesize
    \caption{\textbf{Comparisons to Prior Methods on the EPIC-Kitchen-100 Action Recognition Benchmark.}}
    \label{tab:epic_100}
\setlength{\tabcolsep}{2pt}
\begin{tabular}{|l|ccc|ccc|ccc|ccc|}
\hline
    & \multicolumn{6}{c|}{Overall}                           & \multicolumn{3}{c|}{Unseen Participants} & \multicolumn{3}{c|}{Tail Classes} \\
\cline{2-13}
 Method & \multicolumn{3}{c|}{Top-1 Accuracy} & \multicolumn{3}{c|}{Top-5 Accuracy} & \multicolumn{3}{c|}{Top-1 Accuracy}           & \multicolumn{3}{c|}{Top-1 Accuracy}    \\
\cline{2-13}
                               & \multicolumn{1}{c}{Verb} & \multicolumn{1}{c}{Noun} & \multicolumn{1}{c|}{Action} & \multicolumn{1}{c}{Verb} & \multicolumn{1}{c}{Noun} & \multicolumn{1}{c|}{Action}& \multicolumn{1}{c}{Verb} & \multicolumn{1}{c}{Noun} & \multicolumn{1}{c|}{Action} & \multicolumn{1}{c}{Verb} & \multicolumn{1}{c}{Noun} & \multicolumn{1}{c|}{Action} \\ 
 \hline
                                            TSN~\cite{tsn}                    & 60.18 & 46.03 & 33.19 & 89.59 & 72.90 & 55.13 & 47.42 & 38.03 & 23.47 & 30.45 & 19.37 & 13.88 \\
                                            TRN~\cite{trn}                    & 65.88 & 45.43 & 35.34 & 90.42 & 71.88 & 56.74 & 55.96 & 37.75 & 27.70 & 34.66 & 17.58 & 14.07 \\
                                            TBN~\cite{epic-fusion}                & 66.00 & 47.23 & 36.72 & 90.46 & 73.76 & 57.66 & 59.44 & 38.22 & 29.48 & 39.09 & 24.84 & 19.13 \\
                                            TSM~\cite{lin2019tsm}                     & 67.86 & 49.01 & 38.27 & 90.98 & 74.97 & 60.41 & 58.69 & 39.62 & 29.48 & 36.59 & 23.37 & 17.62 \\
                                            SlowFast~\cite{slowfast} & 65.56 & 50.02 & 38.54 & 90.00 & 75.62 & 58.60 & 56.43 & 41.50 & 29.67 & 36.19 & 23.26 & 18.81 \\
                                            MViT-V2 \cite{MViTv2} & 67.13 & 60.89 & 45.79 & 91.13 & 83.93 & 66.83 & 57.75 & 50.52 & 34.84 & 40.85 & 38.47 & 25.35\\
                                            Ego-Exo \cite{ego-exo} & 66.61 & 59.51 & 44.89 & 91.13 & 82.03 & 65.05 & 56.57 & 48.87 & 33.71 & 40.91 & 38.26 & 25.23 \\
                                            Swin-B \cite{swin} & 67.93 & {58.69} & 46.05 & 90.96 & \textbf{83.77} & 65.23 & 58.69 & \textbf{50.89} & 35.02 & 41.08 & 37.21 & 25.41 \\
                                            \textbf{CVAR (Ours)} & \textbf{69.37} & \textbf{61.03} & \textbf{46.15} & \textbf{91.51} & 81.03 & \textbf{67.05} & \textbf{59.91} & 48.36	& \textbf{35.12} & \textbf{41.93} &	\textbf{38.58} & \textbf{25.99} \\
\hline
\end{tabular}%
\end{table}

\begin{wraptable}{r}{0.3\textwidth}
\centering
\caption{Comparision on NTU RGB+D Action Recognition}\label{tab:ntu_res}
\resizebox{0.3\textwidth}{!}{
\begin{tabular}{|l|c|}
\hline
Method & Top 1  \\
\hline
SlowFast~\cite{slowfast} & 90.22 \\
DANN \cite{ganin2016domain}  & 90.41           \\
MViT-V2 \cite{MViTv2}  & 91.35 \\
Ego-Exo \cite{ego-exo} & 92.14 \\
Swin B \cite{swin}  & 93.72         \\
\textbf{CVAR}   & \textbf{95.93}          \\
\hline
\end{tabular}
}
\end{wraptable}
\noindent
\textbf{Cross-view NTU RGB+D Action Recognition} To further illustrate the effectiveness of our approach when the domain gap is not that large, we conducted an experiment on the NTU RGB+D action recognition dataset.
In this experiment, we use the videos captured from the $0^o$ angle as the source view, while the two other angles ($\pm45^o$) are considered as the target view. 
As shown in Table \ref{tab:ntu_res}, our proposed CVAR has outperformed the other methods by a large margin.  In particular, while the performance of Swin B \cite{swin} achieved 93.72\%, our CVAR approach gained the Top 1 accuracy of 95.95\%. 
These results have shown the effectiveness of our proposed approach in modeling action recognition across views.

\section{Conclusions and Limitations}

\noindent
\textbf{Conclusions}
This paper presents a novel approach for cross-view learning in action recognition (CVAR). 
Using our proposed cross-view self-attention loss, our approach has effectively transferred the knowledge learned from the exocentric to the egocentric view. Moreover, our approach does not require pairs of videos across views, which increases the flexibility of our learning approaches in practice. 
Experimental results on standard egocentric action recognition benchmarks,
have shown our SOTA performance. %
Our method outperforms the prior direct training, pre-training, and domain adaptation methods.

\noindent
\textbf{Limitation of Linear Relation} Modeling the relation in 
Eqn. \eqref{eqn:cross_view_condition} by the linear scale $\alpha$ 
could bring some potential limitations as %
the cross-view correlation of videos and attention maps could be a non-linear proportion and may be subjected to an individual video and its corresponding attention map.
Our future works will consider modeling this relation by a deep network to gain more improvement.

\noindent
\textbf{Limitation of Bounded Distribution Shifts}
Although this assumption allows us to establish the bounded constraint as in Eqn. \eqref{eqn:upper_bound} and further derive into our loss in Eqn. \eqref{eqn:loss_for_unpair}, 
this could also contain some potential limitations. 
If the changes across views of videos (attention maps) are significantly large, this could result in the bounded constraint in  Eqn. \eqref{eqn:upper_bound} is not tight. Thus, the models could not be well generalized w.r.t the large distribution shifts.

\bibliography{egbib}

\end{document}